\documentclass{article}

\PassOptionsToPackage{numbers, compress}{natbib}
\usepackage[preprint]{neurips_2023}

\usepackage[utf8]{inputenc} % allow utf-8 input
\usepackage[T1]{fontenc}    % use 8-bit T1 fonts
\usepackage{hyperref}       % hyperlinks
\usepackage{url}            % simple URL typesetting
\usepackage{booktabs}       % professional-quality tables
\usepackage{amsfonts}       % blackboard math symbols
\usepackage{nicefrac}       % compact symbols for 1/2, etc.
\usepackage{microtype}      % microtypography
\usepackage{xcolor}         % colors
\usepackage{natbib}
\usepackage{graphicx}
\usepackage{bm}
\usepackage{multirow}
\usepackage{musicography}
\usepackage{MnSymbol,wasysym}

\title{May the Dance be with You: 
\\Dance Generation Framework for Non-Humanoids}

\author{%
  Hyemin Ahn\\
  Graduate School of Artificial Intelligence\\
  Ulsan National Institute of Science and Technology \\
  Ulsan, Korea, 44919 \\
  \texttt{hyemin.ahn@unist.ac.kr} \\
}

\begin{document}

\maketitle

\begin{abstract}
    We hypothesize dance as a motion that forms a visual rhythm from music, where the visual rhythm can be perceived from an optical flow.
    If an agent can recognize the relationship between visual rhythm and music, it will be able to dance by generating a motion to create a visual rhythm that matches the music.
    Based on this, we propose a framework for any kind of \textit{non-humanoid agents} to learn how to dance from \textit{human videos}.
    Our framework works in two processes: (1) training a reward model which perceives the relationship between optical flow (visual rhythm) and music from human dance videos, (2) training the non-humanoid dancer based on that reward model, and reinforcement learning. 
    Our reward model consists of two feature encoders for optical flow and music. They are trained based on contrastive learning which makes the higher similarity between concurrent optical flow and music features.
    With this reward model, the agent learns dancing by getting a higher reward when its action creates an optical flow whose feature has a higher similarity with the given music feature.
    Experiment results show that generated dance motion can align with the music beat properly, and user study result indicates that our framework is more preferred by humans compared to the baselines.
    To the best of our knowledge, our work of non-humanoid agents which learn dance from human videos is unprecedented. An example video can be found at \url{https://youtu.be/dOUPvo-O3QY}.
\end{abstract}

\section{Introduction}

\textit{``What is dance?"}---though there exist various answers to this question \cite{copeland1983dance}, what many people would agree is that dance, which can be interpreted as physical, social, and cultural behavior, is one of the unique properties of humans \cite{hanna1987dance}. 
Perhaps for this reason, most researchers of AI-based dance generation consider their agent or robot dancers would have a human-like upper body or full body \cite{creative2017, tang2018dance, ahn2020generative, li2021ai, siyao2022bailando, tseng2022edge, boukheddimi2022robot}. 
But does an agent or a robot have to resemble a human to be able to dance?
The answer would be `No'---as Boston Dynamics has already shown from their videos that non-humanoid robot is also able to dance perfectly \cite{spotsonit}. One regrettable thing would be that their dance is manually designed by humans \cite{alltogethernow}.
Then, what about letting a non-humanoid agent learn how to dance from data?
If we rely on supervised learning, it would be possible to train non-humanoid agents how to dance after we go through a very tedious and time-consuming data collection process.

In this paper, our goal is to circumvent this data collection process and train any kinds of non-humanoid agents how to dance. 
To this end, we propose a dance generation framework, which trains \textit{non-humanoid dancers from human dance videos}. 
To realize this, we set one hypothesis: \textit{dance is a movement that forms a visual rhythm from music, and the visual rhythm can be recognized from an optical flow}. Note that our hypothesis is not unprecedented, rather, it has been inspired by a previous study \cite{yu2022self} of music-dance representation learning.
With this hypothesis, if the optical flow created by the agent's musical movement has a similar pattern to the one from a human, we claim that the agent's movement can also be recognized as a dance.

Our non-humanoid dancers are trained in two processes.
First, we train a reward model which learns the relationship between optical flow (visual rhythm) and music from the AIST dance video database \cite{tsuchida2019aist}. The reward model consists of two feature encoders, one for optical flow and another for music. They are trained by contrastive learning framework as \cite{chen2020simple}, such that making the higher similarity between features from concurrent optical flow and music.
Second, based on the trained reward model, our framework uses reinforcement learning (RL) algorithms \cite{ppo, sac} to train non-humanoid dancers.
During this training, the agent obtains a higher reward if its action creates an optical flow, whose feature has a higher similarity with the given music feature.

In experiments, we include non-humanoid agents such as Cartpole and the one-armed robot UR5. In a qualitative study, we present the non-humanoid dances generated from music that would be familiar to readers, showing how the agent's movement matches the given music.
In a quantitative study, we compare our framework with a simple baseline based on various metrics including beat align scores \cite{li2021ai}, as well as user study results. 
To the best of our knowledge, our work is the first learning-based approach to make any kind of non-humanoid agents dance from human dance videos.

\section{Related Work}
\subsection{Dance Generation for Human-like Agents}
Studies of dance generation for avatars or agents in a human form have been developed rapidly thanks to the advances in deep neural networks and generative models.
Since dance is a sequence of poses generated from an audio signal, deep neural networks such as LSTM \cite{lstm} or Transformer 
\cite{transformer} that can handle sequential data are often used for dance generation \cite{tang2018dance, li2021ai, siyao2022bailando}.
Also, there are other works that proposed auto-regressive models consisting of one-dimensional convolution layers for generating 2D \cite{lee2018listen} or 3D \cite{ahn2020generative} dance motion.
In addition, with the advent of the diffusion models \cite{ddpm}, there exists a model capable of editable dance generation \cite{tseng2022edge}, which is state-of-the-art until now, to the best of our knowledge.

Many of these works have been facilitated by the emergence of new large-scale data called AIST dance video database \cite{tsuchida2019aist} and AIST++ dance motion dataset \cite{li2021ai}.
The AIST dance video database provides 1,389 human dance videos from 60 music pieces of 10 genres.
% From this database, 3D human keypoint annotations as well as camera parameters can be constructed, and they are provided from the
Based on multi-camera videos in this database, the AIST++ dance motion dataset extracts 3D human keypoint annotations as well as camera parameters and provides 3D human dance motion which can be visualized with SMPL mesh models \cite{SMPL:2015}.
Most of the existing works on human characters' dance motion generation rely on this AIST++ dataset.
In this paper, we employ the AIST dance video database to train non-humanoid agents how to dance.

Not only the human-like agents in the simulator, but also there exist studies of dancing real-world humanoid robots.
To do this, some studies collect the robot posture library, analyze the music (i.e., beat tracking), then use reinforcement learning \cite{kumra2015dual} or optimal control strategy \cite{boukheddimi2022robot} to plan a trajectory that maps current posture to next posture. Others tried to use a generative model such as variational autoencoders (VAEs) \cite{VAE}, such that creative robot dance motion can be generated \cite{creative2017}.
Note that these studies employ the upper body of humanoids, since planning a stable full-body motion is challenging.
Of course, for a full-body dance motion of humanoids, there is a fancy video from \cite{doyouloveme}.

\subsection{Dance Generation for Non-Humanoid Robots}
Researchers on social robots have been interested in non-humanoid robots that can dance in front of humans.
For instance, a social robot named Keepon \cite{keepon}, which is designed for rhythmic interaction with children, has only a head and belly but is able to dance. With 4 degrees of freedom, Keepon can synchronize its motion to the rhythm perceived from the surrounding environment \cite{keepondance1}. 
There is another robot named Pleo, which looks like a dinosaur and is designed for human-robot interaction through dance \cite{curtis2011dance}.
It can learn new dance moves from human demonstrations, which are represented as trajectories of an object held and moved by the human hand.
We believe that these robots would have contributed to helping children to become familiar with robots.
However, there are disadvantages that the robots would only perform certain movements to the rhythm, or a human user needs to teach robots directly how to dance.

In addition to these social robots focused on interaction with humans, there are several works focused more on planning trajectories so that non-humanoid robots can appear to be dancing.
For instance,
\cite{schollig2010synchronizing}
proposed a control system for flying vehicles such as quadrocopters to perform side-to-side movements synchronized with the rhythm of given music.
Not only the flying robots, but also there exist a legged robot that is capable of real-time dancing \cite{bi2018real}.
As their robot extracts the beat in real-time, it starts dancing by arranging motions from their predefined motion library. This arrangement is processed by Markov chain-based probabilistic model, which depends on the previous motion and the detected music tempo.
These studies made flying or legged robots look like dancing to the rhythm. 
However, they focus more on how to use manually designed motion libraries by synchronizing them with rhythm, which implies that constructing a custom motion library is inevitable for their systems.
In contrast, our dance generation framework does not need any pre-designed motions, rather it uses only human dance videos and teaches any kind of non-humanoid agents to dance by imitating how humans form a visual rhythm synchronized with music.

\section{Methodology}

\begin{figure}[t]
\centering
\includegraphics[width=0.8\columnwidth]{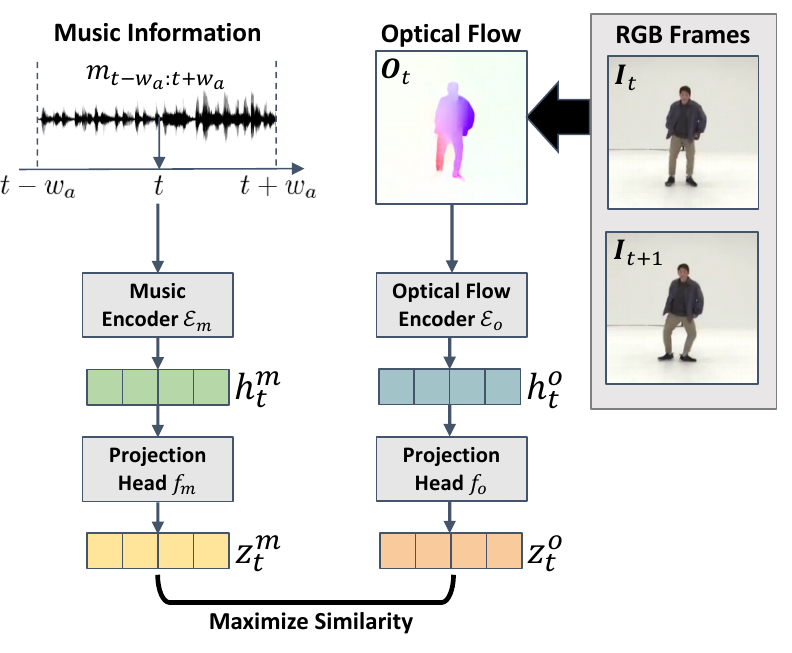}
\caption{Illustration of how the proposed reward model consisting of two encoders and two projection heads are trained based on contrastive learning.}
\label{fig:reward}
\end{figure}

\subsection{Training Reward Model}\label{sec:reward}

Let $m_t \in \mathbb{R}^{d_m}$ denote a raw music feature at time $t$.
% , extracted from a toolbox named \textit{librosa} \cite{mcfee2015librosa}. 
% Details of how we extract raw music feature $m_t$ from audio data are explained in Section~\ref{sec:imp}.
And let $a_t$ denote the agent's action at time $t$.
Our objective is to make the agent plan its action to dance.
When a set of raw music features $\mathcal{M}=\{m_1,\ldots,m_T\}$ is given, the generated dance would be represented as $\mathcal{A} =\{a_1,\ldots a_T\}$.

In order to teach an agent based on reinforcement learning, a reward function needs to be defined. 
Therefore, we first propose a reward model based on our main hypothesis: \textit{dance is a movement that forms a visual rhythm from music, and the visual rhythm can be recognized from an optical flow.}
Our reward model is designed to return a higher value if the visual rhythm (optical flow) created by the agent's action is more correlated with the given music. 
To this end, we employ contrastive learning \cite{chen2020simple} to train $\{\mathcal{E}_{o}(\cdot), \mathcal{E}_{m}(\cdot)\}$, which are two encoders for optical flow and music, respectively.
Since our goal is to learn the relationship between visual rhythm (optical flow) and music in human dance, we train our encoders, which are components of the proposed reward model, based on the AIST dance video database \cite{tsuchida2019aist}.

Figure \ref{fig:reward} shows an illustration of how we train our reward model.
First, an optical flow $\bm{O}_t \in \mathbb{R}^{2\times w\times h}$ is extracted from two consecutive images $\bm{I}_{t-1}, \bm{I}_{t} \in \mathbb{R}^{3\times w\times h}$.
% How we extract the optical flow is mentioned in Section~\ref{sec:imp}.
% based on the RAFT model \cite{teed2020raft}.
Second, a set of raw music features inside a time window $m_{t-w_a:t+w_a}=\{m_{t-w_a},\ldots, m_{t+w_a}\} \in \mathbb{R}^{(2w_a+1) \times d_m}$ is chosen as an input to music encoder $\mathcal{E}_m$. 
Third, $m_{t-w_a:t+w_a}$ and $\bm{O}_t$ are given to each encoder $\mathcal{E}_m$ and $\mathcal{E}_o$, such that representations of music and optical flow can be obtained.
We denote each representation vector at time $t$ as $h_t^m=\mathcal{E}_m(m_{t-w_a:t+w_a})$ and $h_t^o=\mathcal{E}_o(\bm{O}_t)$.
The representation vectors are given to projection heads, resulting in $z_t^o=f_o(h_t^o)$ and $z_t^m=f_m(h_t^m)$.
Finally, whole networks are trained by maximizing the similarity between $z_t^o$ and $z_t^m$, based on the infoNCE loss as proposed in \cite{oord2018representation}.

\subsection{Training Non-Humanoid Dancer}

\begin{figure}[t]
\centering
\includegraphics[width=\columnwidth]{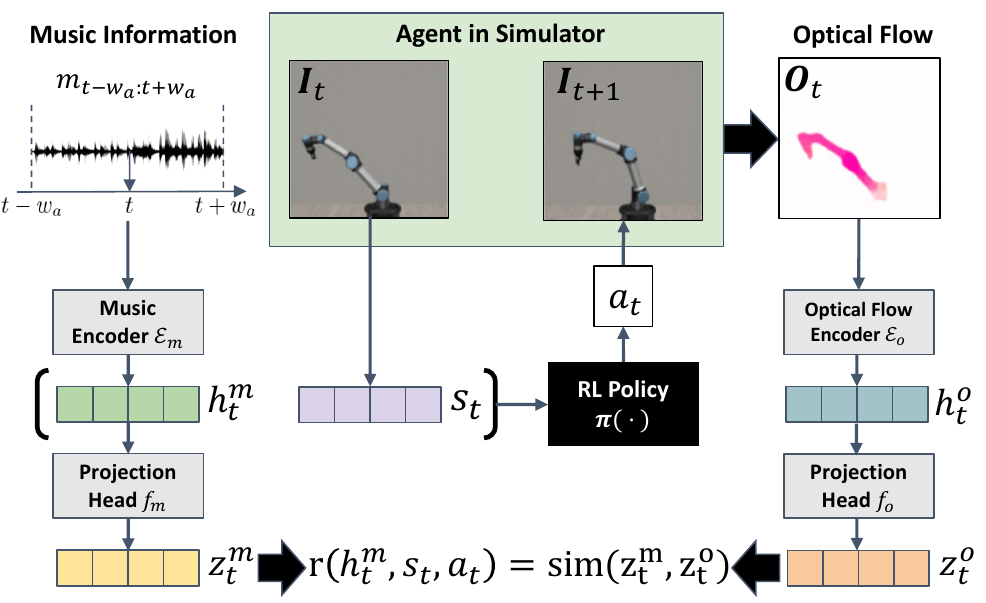}
\caption{Illustration of how the proposed framework teaches the non-humanoid dancer based on reinforcement learning (RL). \texttt{sim($\cdot, \cdot$)} denotes the cosine similarity between two input vectors. 
% \textcolor[rgb]{0, 0, 1}{add pi to figure! make figure to pdf!}
}
\label{fig:rl}
\end{figure}

Once the reward model is trained, the reward model can be universally used to teach any kind of non-humanoid agent how to dance. Figure \ref{fig:rl} shows how the trained reward model is used to train a non-humanoid agent how to dance based on reinforcement learning.
Detailed and ordered descriptions are as follows:
\begin{itemize}
    \item[1.] A set of raw music features $m_{t-w_a:t+w_a}$ passes through the trained representation encoder $\mathcal{E}_m$ and projection head $f_m$, and music representation vector $h_t^m$ and $z^m_t$ are obtained. 
    \item[2.] $h_t^m$ is concatenated with the state vector $s_t$ (i.e., position, velocity) of the agent, and given to the policy function $\pi(\cdot)$ to obtain action $a_t = \pi([h_t^m; s_t])$.
    \item[3.] After the agent conducts the action $a_t$ and moves to the next state on time $t+1$, an optical flow $\bm{O}_t$ is obtained from consecutive images $\bm{I}_t, \bm{I}_{t+1}$ observing the agent.
    \item[4.] The optical flow $\bm{O}_t$ passes through its trained representation encoder $\mathcal{E}_o$ and projection head $f_o$, obtaining optical flow representation vector $h_t^o$ and $z_t^o$.
    \item[5.] Obtain a reward $r(h_t^m, s_t, a_t)$ by calculating the cosine similarity between $z_t^m$ and $z_t^o$, and use it for training the reinforcement learning algorithms.
\end{itemize}

After the agent finishes its training based on the reinforcement learning algorithms such as PPO \cite{ppo} and SAC \cite{sac}, the final dance motion $\mathcal{A} =\{a_1,\ldots a_T\}$ is obtained by iterating processes 1 to 3 mentioned above.

% However, there could be an argument that our trained reward model would not be necessary to teach the non-humanoid dancers.
% Instead, the distance between the optical flows from agents and humans can be used to calculate the reward. 
% For instance, if there is a human dance video, its audio can be given as input to the agent, and the optical flow formed from the agent's action can be compared with the real optical flow from the original human video.
% Regarding this, our comparison study result is in Section \ref{sec:exp}, showing it is better to use the reward model that has been trained based on a large-scale human dance video database.

\subsection{Implementation Details}\label{sec:imp}
\paragraph{Optical Flow Extraction} We first extract optical flows from the AIST dance video database \cite{tsuchida2019aist} using the RAFT model \cite{teed2020raft} at 60 FPS. 
Among various dance video types (i.e., basic dance, advanced dance, group dance) in the AIST dance video database, we choose 1,200 videos from basic dance.
Note that the AIST dance video database contains a set of videos that simultaneously film a dancer from multiple camera views.
Among various multi-view videos, we choose the video that films a human dancer from the front view. 

\paragraph{Music Feature Extraction} When extracting $m_t$, we use \textit{librosa} as well as the feature extraction strategy from \cite{li2021ai}, resulting in a 35-dimensional feature including envelope, MFCC, chroma, one-hot peaks, and beats information.
This audio information is also processed at 60 FPS, such that it can be synchronized with extracted optical flows.

\paragraph{Reward Model Training} An optical flow encoder $\mathcal{E}_o$ is based on ResNet-50 \cite{resnet} with the first convolutional layer being modified to process an optical flow, which has two input channels. When training the reward model, a data augmentation process is conducted by randomly cropping the optical flow $\bm{O}_t$ and resizing it to $224 \times 224$, and giving it to the optical flow encoder $\mathcal{E}_o$.
A music encoder $\mathcal{E}_m$ is a Transformer encoder \cite{transformer}, whose input $m_{t-w_a:t+w_a} \in \mathbb{R}^{(2w_a+1) \times d_m}$ is from a symmetric time window $(t-w_a) \sim (t+w_a)$.
Let $o_{t-w_a:t+w_a} \in \mathbb{R}^{(2w_a+1) \times d_o}$ denote the output from this music encoder $\mathcal{E}_m$. Then, the center output $o_t$ is used as the representation vector $h_t^m$.
The two projection heads $f_o$ and $f_m$ are multi-layer perceptrons consisting of two fully connected layers with a GELU layer in the middle.
Finally, the obtained representations $h_t^o, z_t^o, h_t^m, z_t^m$ are 512-dimensional vectors.
To train the reward model, we used a single RTX 3090 GPU for around 7 days.

\paragraph{Non-Humanoid Dancer Training} For training non-humanoid dancers with reinforcement learning, we use Proximal Policy Optimization (PPO) \cite{ppo} for agents with discrete action space (i.e,. CartPole), and Soft Actor-Critic (SAC) \cite{sac} for agents with continuous action space (i.e., UR5 Robot). Note that the parameters of $\mathcal{E}_m$ and $\mathcal{E}_o$ are frozen when the pre-trained reward model is used. More details including the hyperparameter information can be found in our supplementary material. 

\section{Experiment} \label{sec:exp}
\subsection{Agents and Simulator}
For non-humanoid dancers, we choose two agents from different simulators: CartPole in Gym, and one-armed robot UR5 with gripper in RoboSuite \cite{zhu2020robosuite}.
We selected these agents considering the degree of complexity in their observation and action space: from simple (CartPole) to complicated (UR5).
Note that agents that resemble humans are excluded from our experiments since our goal is to train non-humanoid dancers. 

When training the agent with reinforcement learning, not only the reward but also an additional penalty is applied for special cases.
For the CartPole agent, the pole can be moved freely without being constrained to remain vertical. Instead, if the agent moves out of camera view, the episode ends and a penalty of -1 is given. If the agent is further away from the center of the camera view, a smaller reward will be given.
For the UR5 agent, if its body moves lower than its base position, a penalty of -1 is given for moving too far down. It is also possible to give a penalty when self-collision happens from UR5's body. However, we exclude the self-collision penalty from the experiment because it is computationally expensive to check self-collision at every step. 

\begin{figure}[t]
\centering
\centerline{\includegraphics[width=0.95\columnwidth]{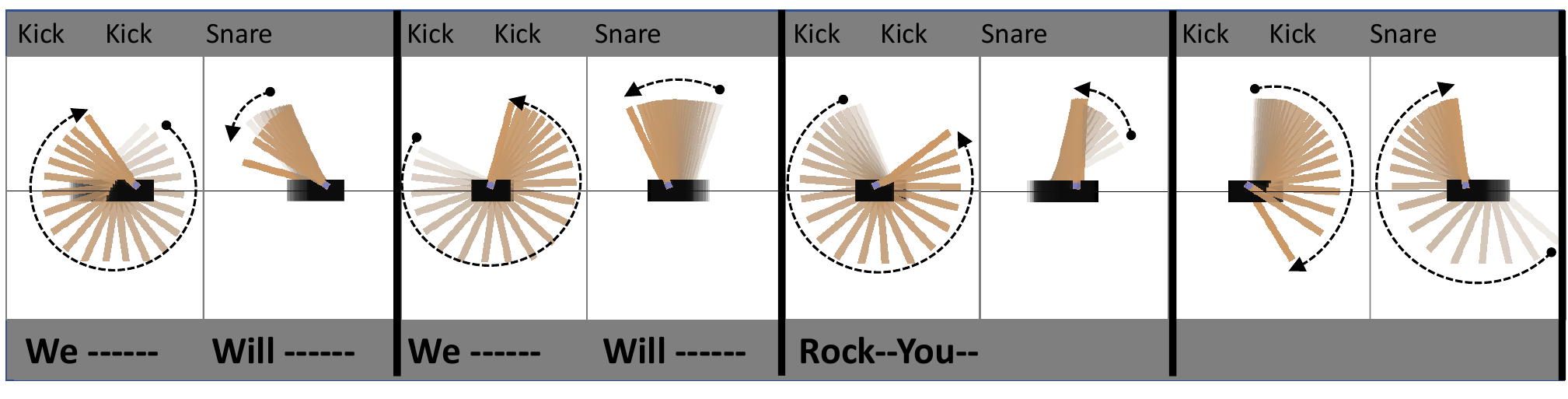}}
\caption{Generated dance of CartPole agent with a song ``We will rock you'' by Queen.}
\label{fig:cartpolerock}
\end{figure}

\begin{figure}[t]
\centering
\centerline{\includegraphics[width=0.95\columnwidth]{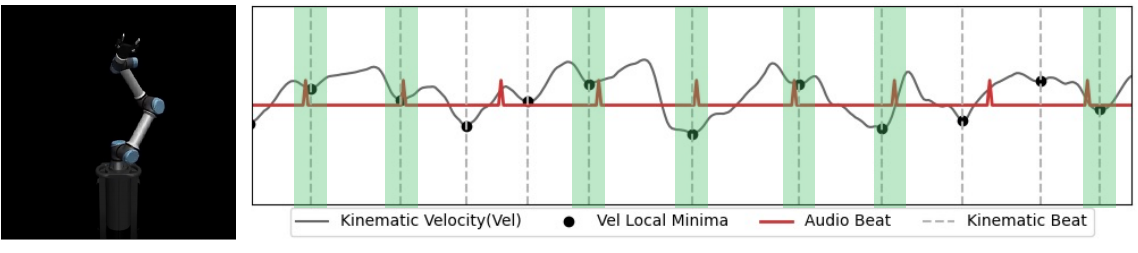}}
\caption{Example of how the kinematic beat aligns with the music beat when UR5 agent dances.}
\label{fig:ur}
\end{figure}

\subsection{Qualitative Result}
Figure \ref{fig:cartpolerock} shows the visualization of the generated dance of our CartPole agent based on the song ``We will rock you'' by Queen.
% It is shown that the trained CartPole agent makes its movements bigger or smaller according to the audio dynamics as well as the main beat of the song.
In the chorus of this song, tension is created by playing the kick drum twice. Then, a snare is played once and that tension is relieved.
In the generated dance, the movement becomes larger when the tension is higher, and becomes smaller when tension is lower.
This creates a visual rhythm that matches the music, making humans feel that the CartPole is dancing.
For better visualization, we highly recommend readers to watch our non-humanoid dance videos in our supplementary materials.

Figure~\ref{fig:ur} shows how the kinematic beat aligns with the music beat when the UR5 agent dances.
This dance is generated based on the music sample (ID: mJB1) from the AIST video database. 
In this figure, the red line indicates the music beat information which is detected from the \textit{librosa} \cite{mcfee2015librosa} package.
The gray line is the norm value of the joint velocity vector of UR5 and represents the kinematic velocity of the agent. 
When the agent's movement is relatively slow so that the corresponding velocity has a local minimum value, we denote that the kinematic beat, which is indicated with a vertical gray dotted line, is formed at that moment.
This example shows that the kinematic beat is well aligned with the beat of the given music. 
Of course, it is not desirable to evaluate the movement as a good dance if it matches only to the music beat.
But still, this result is somewhat in line with our main hypothesis, ``dance is a motion that creates a visual rhythm to the music'', since the kinematic beat is correlated with the visual rhythm.
Related quantitative results and discussion will be shown in the next section. 
% 이 그림에서, 가로축은 시간을 의미하며, 빨간선은 librosa 패키지에서 검출된 오디오 음악의 비트 정보를 나타낸다. 회색 선은 UR5가 가진 joint velocity vector의 norm값으로, agent의 kinematic velocity를 나타낸다. agent의 움직임이 상대적으로 느려져 해당 velocity의 값이 local minima값을 가질 때, 우리는 그 시점에 visual rhythm을 구성하는 beat이 형성된다고 이야기 하며, 이는 vertical한 회색 점선으로 그림에 표시되어 있다. 그림에서 보여지는 예시는 agent가 형성하는 visual rhythm의 beat가 주어진 음악의 beat에 잘 align되어있음을 보여준다. 물론, 음악의 beat 정보에 잘 맞춰서 만든 움직임이 좋은 춤이라고 평가하기는 힘들지만, 우리의 가정인 "춤은 음악에 맞춰 비쥬얼 리듬을 만드는 모션이다" 에 어느정도 부합하는 결과라고 우리는 생각한다. 이와 관련된 quantitative 수치 값들은 다음 섹션에 소개되고 discuss된다. 

\subsection{Quantitative Result}
\subsubsection{Comparison Baselines}
\paragraph{BPM-based Control}
The proposed framework is suggested under our hypothesis that dance is a motion creating a visual rhythm that matches the music. Based on this hypothesis, another simpler baseline can be defined, which is to create the motion by changing the kinematic velocity of the agent body in synchronization with the quarter note ( \musQuarter \, ).
The BPM-based Control in Table~\ref{tab:stats} refers to this baseline. In the case of UR5, the joint angle velocity control vector is randomly changed at each quarter note beat to create a motion that appears to be synchronized with the beat. 
In the case of CartPole, this baseline cannot be used since the agent is not able to remain inside the camera view if its action randomly changes.

\begin{figure}[t]
\centering
\centerline{\includegraphics[width=0.9\columnwidth]{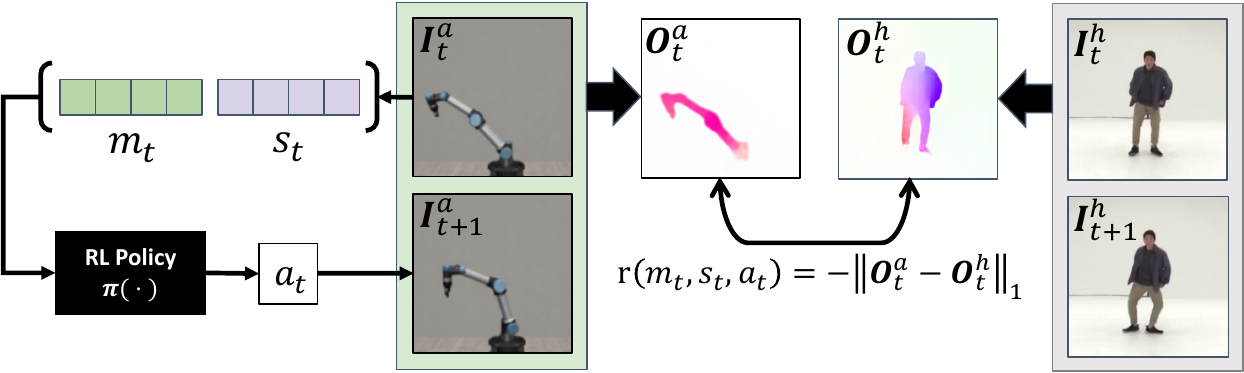}}
\caption{Illustration of how the proposed framework without the reward model can teach the non-humanoid dancer. Instead of a pre-trained reward model, it is also possible to set the reward as an L1-norm distance between optical flows from the agent and human.}
\label{fig:baseline}
\end{figure}

\paragraph{Proposed Framework without Reward Model}
Based on our main hypothesis, a simplified version of the proposed framework can be also suggested, which is shown in Figure~\ref{fig:baseline}. We refer to this baseline as the Proposed without Reward Model (RM) in Table~\ref{tab:stats}. This baseline framework does not employ the pre-trained reward model consisting of encoders and projection heads for optical flow and music. 
Rather, it first gives the raw music feature $m_t$ as well as the agent state $s_t$ to the reinforcement learning policy $\pi(\cdot)$, obtains the agent action $a_t$.
After the agent moves with $a_t$, the optical flow $\bm{O}_t^a$ can be obtained from the agent. 
This optical flow $\bm{O}_t^a$ can be compared with the optical flow of human $\bm{O}_t^h$, which can be directly obtained from the human dance videos with the same audio of $m_t$. Finally, we define the simplified reward with the L1-norm distance between two optical flows, such that $r_t(m_t, s_t, a_t) = -\|\bm{O}_t^a - \bm{O}_t^h \|_1$.

\subsubsection{Motion-Music Correlation}
\paragraph{Beat Calculation}
To check the correlation between the generated dance motion and music, it is necessary to calculate the beat information. 
To do this, we follow the strategy as existing work \cite{li2021ai} suggests.
Let $B^k=\{t_i^k\}$ denote the kinematic beat, which is extracted from the local minima of the kinematic velocity. Note that the kinematic beat is highly correlated with the visual rhythm. Here, $t_i^k$ denotes the time frame of $i$-th kinematic beat.
Let $B^b=\{t_j^b\}$ denote the music beat, which can be extracted by a python package \textit{librosa} \cite{mcfee2015librosa}. 
Here, $t_j^b$ denotes the time frame of $j$-th music beat. Similarly, audio peak information $B^p=\{t_j^p\}$ can also be obtained.
Based on $B^k$, $B^b$, and $B^p$, we calculate the motion-music correlation by obtaining alignment scores as well as F1@(\musSixteenth, \musThirtySecond, \musSixtyFourth) scores.

\paragraph{Alignment Scores}
A beat alignment score is a metric for measuring motion-music correlation, proposed by \cite{li2021ai}.
It is defined as the below equation, which obtains the average time distance between every kinematic beat and its nearest music beat:
\begin{equation} \label{eqn:align}
\textrm{BeatAlign}=\frac{1}{N} \sum_{i=1}^N \exp \big( -\frac{\min_{\forall t_j^b \in B^b }\|t_i^k-t_j^b\|^2}{2\sigma^2}\big) 
\end{equation} 
Here, we set $\sigma=3$ in our experiments, considering the 60 FPS of motion and music as \cite{li2021ai} did. 
In addition, we also calculate the peak alignment score, which is the average distance between every kinematic beat $B^k$ and its nearest audio peak $B^p$, by using the equation (\ref{eqn:align}) with $B^p$ instead of $B^b$.

The result of the beat and peak alignment scores are shown in Table~\ref{tab:stats}. Note that numbers are obtained from dances generated based on the music in AIST video dataset. In the case of UR5, it is shown that BPM-based control obtains the highest scores. We speculate this would be due to its control frequency based on the BPM information, which is highly correlated with the music beat and peak information.
When comparing the proposed framework with and without the reward model, it is shown that the proposed framework with the reward model performs better in general. 
However, one might argue that there is no significant difference between the alignment scores. 
% When there is a reward model, the beat alignment score decreases slightly, but the peak alignment score increases instead.
This would be because the proposed framework without the reward model can learn how to create a visual rhythm to music, since it is also designed based on our main hypothesis. 
Despite this, we still believe that the proposed framework with the reward model can learn to dance better, since the reward model will be able to comprehend the music by considering various musical factors, not just the beats or peaks of audio. This claim can be validated by the user study results in the next section, where the proposed framework with the reward model outperforms other baselines. 

\paragraph{F1@(\musSixteenth, \musThirtySecond, \musSixtyFourth) \ Scores}
Alignment scores indicate how close the kinematic beats are to the beats or peaks of the audio. 
However, since it is more of a calculation related to precision, it does not include information related to the recall, such as how much of the entire beats or peaks of the audio match kinematic beats. Therefore, we suggest a new metric, which is F1@(\musSixteenth, \musThirtySecond, \musSixtyFourth) score. Here, the notes \musSixteenth, \musThirtySecond, \musSixtyFourth \, denote the thresholds for judging true positives. For instance, when the threshold is set to \musSixteenth, it is defined as true positive if the time distance between the kinematic beat and its closest music beat or peak is smaller than the length of \musSixteenth.

The result of beat and peak F1 scores are shown in Table~\ref{tab:stats}. 
In the case of UR5, it is shown that the BPM-based control shows the highest performance in Beat F1 scores. However, its performance in Peak F1 scores is the lowest, which is contrary to having a high peak alignment score. 
Note again that BPM-based Control moves the agent according to the quarter note beat. This makes the visual rhythm more likely to be formed closer to the music beat or peak, increasing the alignment scores which are calculated in terms of precision. 
Compared to this, our F1 scores also consider a recall, which obtains how much of the entire audio beats or peaks match the kinematic beat. 
If the agent moves according to the BPM, it can cover most of the audio beats represented by quarter notes, so the Beat F1 score is also high.
However, its Peak F1 score becomes lower since the agent can miss quite a few of the audio peaks, which is correlated to the audio beats but occurs more frequently. 

\begin{table}[]
\centering
\caption{Comparison results between the proposed method with the reward model (RM) and baselines.}
\begin{tabular}{@{}llllccccc@{}}
                              &                                                 &                                                &      & \multicolumn{4}{c}{\textbf{Motion-Music Correlation}} & \textbf{User Study}              \\ \midrule
                              &                                                 &                                                &      & \textbf{F1@\musSixteenth$\uparrow$ }  & \textbf{F1@\musThirtySecond$\uparrow$}   & \textbf{F1@\musSixtyFourth$\uparrow$ }  & \textbf{AlignScore$\uparrow$}   & \textbf{LoseRate$\downarrow$}                \\ \midrule
\multicolumn{3}{l}{\multirow{2}{*}{\textbf{AIST++}}}                                                                                        & \textbf{Beat} & 0.269     & 0.164     & 0.087     & 0.259        & \multirow{2}{*}{-}      \\
\multicolumn{3}{l}{}                                                                                                             & \textbf{Peak} & 0.290     & 0.173     & 0.087     & 0.588        &                         \\ \midrule
\multirow{6}{*}{\textbf{UR}}           & \multicolumn{2}{l}{\multirow{2}{*}{\begin{tabular}[c]{@{}l@{}}\textbf{BPM-based}\\ \textbf{Control}\end{tabular}}} & \textbf{Beat} & 0.612     & 0.353     & 0.175     & 0.247        & \multirow{2}{*}{69.3\%} \\
                              & \multicolumn{2}{l}{}                                                                             & \textbf{Peak} & 0.766     & 0.463     & 0.233     & 0.545        &                         \\ \cmidrule(l){2-9} 
                              & \multicolumn{2}{l}{\multirow{2}{*}{\begin{tabular}[c]{@{}l@{}}\textbf{Proposed}\\ \textbf{w/o RM}\end{tabular}}}   & \textbf{Beat} & 0.586     & 0.311     & 0.142     & 0.210        & \multirow{2}{*}{63.3\%} \\
                              & \multicolumn{2}{l}{}                                                                             & \textbf{Peak} & 0.807     & 0.499     & 0.247     & 0.539        &                         \\ \cmidrule(l){2-9} 
                              & \multicolumn{2}{l}{\multirow{2}{*}{\begin{tabular}[c]{@{}l@{}}\textbf{Proposed}\\ \textbf{w/ RM}\end{tabular}}}    & \textbf{Beat} & 0.571     & 0.304     & 0.152     & 0.208        & \multirow{2}{*}{33.6\%} \\
                              & \multicolumn{2}{l}{}                                                                             & \textbf{Peak} & 0.806     & 0.499     & 0.248     & 0.542        &                         \\ \midrule
\multirow{4}{*}{\textbf{CartPole}}     & \multicolumn{2}{l}{\multirow{2}{*}{\begin{tabular}[c]{@{}l@{}}\textbf{Proposed}\\ \textbf{w/o RM}\end{tabular}}}   & \textbf{Beat} & 0.400     & 0.181     & 0.073     & 0.184        & \multirow{2}{*}{77.3\%} \\
                              & \multicolumn{2}{l}{}                                                                             & \textbf{Peak} & 0.505     & 0.271     & 0.122     & 0.517        &                         \\ \cmidrule(l){2-9} 
                              & \multicolumn{2}{l}{\multirow{2}{*}{\begin{tabular}[c]{@{}l@{}}\textbf{Proposed}\\ \textbf{w/ RM}\end{tabular}}}    & \textbf{Beat} & 0.395     & 0.209     & 0.100     & 0.206        & \multirow{2}{*}{22.6\%} \\
                              & \multicolumn{2}{l}{}                                                                             & \textbf{Peak} & 0.503     & 0.290     & 0.137     & 0.536        &                         \\ \bottomrule
\end{tabular}
\label{tab:stats}
\vspace{-4.5mm}
\end{table}

When comparing the proposed framework with and without the reward model, it is shown that the F1 score of the one with the reward model outperforms more as the note-based threshold becomes shorter.
In the case of UR5, there is not much difference in F1 scores between the two frameworks.
But in the case of CartPole, it is shown that the difference in F1 and Alignment Scores between the two frameworks is greater. 
This is due to the disadvantage that the episode length is limited during reinforcement learning if the pre-trained reward model does not exist. 
As mentioned, our framework without the reward model obtains the reward from the L1-norm distance between two optical flows from the ground truth human video and the agent in a simulator.
This implies that the agent without the reward model can obtain the reward only if the ground truth human dance video is being played.
However, the length of `basic' dance videos in the AIST database is short ($\sim$10 seconds), such that the agent without a reward model cannot interact with the environment for a longer time.
Such a short episode length is not enough to teach dancing to a CartPole agent, which should dance while staying inside the camera view at the same time. 
On the contrary, if there is a pre-trained reward model, one can train the agent without the limitation in episode length, such that the agent can learn with plenty of time how to stably dance during the whole music. 
We believe this makes the F1 score and alignment score higher for the framework with the reward model than the framework without the reward model.
This main advantage of our framework will apply not only to the CartPole, but also to agents that should dance while satisfying the condition to prevent the episode from being halted.

%  이는 REWARD MODEL이 없는 제안하는 방법론이 학습에 사용하는 에피소드의 길이가 한정적 이라는 단점 때문이라고 생각된다. 앞서 언급된 것처럼, 리워드 모델이 없는 방법론은 사람의 비디오에서 얻어지는 옵티컬 플로우와 에이전트로부터 얻어지는 옵티컬 플로우 사이의 l1 norm distance를 통해 리워드를 계산해 낸다. 하지만 aist dance video의 basic dance 비디오들의 길이는 약 30초 되는 원본 음악보다는 짧은 10초 정도여서 그 기간 동안만 사람의 옵티컬 플로우를 얻을 수 있다. 즉, 리워드 모델이 없는 프레임워크 내에서 학습을 위해 움직이는 agent는 오랫동안 환경과 상호작용하면서 리워드를 얻어낼 수가 없는 것이다.  하지만 카트폴이 화면 안에 존재함과 동시에 춤을 잘 추도록 만들려면, AGENT의 행동이 FAILURE라고 결정되는 경우가 없는 UR5와는 다르게, CARTPOLE의 경우 화면 밖으로 나가버리면 EPISODE를 HALT하게 되니, 더 긴 시간동안의 에피소드에 기반해 학습을 진행해 오랫동안 화면 안에서 균형을 잡으며 춤을 추는 법을 배울 필요가 있는데, 리워드가 없는 프레임 워크는 그것이 불가능 하니 카트폴에 대한 학습 자체가 잘 이루어지지 않은 것이다. 이와 반대로, 사전에 학습한 리워드 모델을 가지고 있으면 에피소드의 길이에 구애 받지 않고 원하는 시간동안 트레이닝을 진행할 수 있기 때문에 f1 score 뿐만 아니라 alignment score에서도 높은 성능을 보인것이다. 

% 학습에 사용하는 에피소드의 길이에 제한이 있기 때문에, 

\subsubsection{User Study}
We conducted a user study to validate whether humans prefer the dance moves generated by the proposed framework. 
We recruited 50 human subjects through a platform called Prolific. Each subject was asked to watch nine pairs of 30-second dances, and to answer which non-humanoid agent dances better to the given music.
Here, the comparison videos were paired between the dances created by the proposed framework and the ones created by the baselines.
Detailed information such as which music pieces were used, and in what order the dances were paired and shown to the users, are in our supplementary material. Note that our user study methodology referenced \cite{li2021ai}.

The result of the user study is shown as a LoseRate in Table~\ref{tab:stats}. A lower LoseRate value indicates that the generated dance was more preferred by human subjects when compared with others. For instance, in the case of cartpole, it shows that 77.3\% of dance motions from the BPM-based control approach was not preferred by humans when compared with the proposed framework with the reward model. 
Overall results show that the dance moves from the proposed framework with the reward model were preferred over other baselines.
In particular, in the case of UR5 which has a larger degree of freedom than CartPole, the proposed framework is preferred with a larger deviation.
% , showing that the higher motion-music correlation can result in higher human preference. 
% It would be because the factor evaluating the dance quality is not necessarily whether the motion aligns well with the music beat or peak. Related discussion is more in the next section of Limitation. 

% 우리는 제안하는 방법론이 생성하는 춤 동작들이 사람들로부터 더 선호받을 수 있는지를 확인하기 위해 유저 스터디를 진행하였다. 
% 우리는 prolific이라는 플랫폼을 통해 총 50명의 휴먼 서브젝트들을 섭외하였다. 그리고, 각 서브젝트들에게 15분 동안 30초 남짓의 춤들의 한 쌍을 9개 보여줬으며, 둘 중 어떤 춤이 가장 주어진 음악에 잘 맞춰 춤을 추었는지 선택해 달라고 하였다. 이 때 제안하는 방법론이 만든 춤과 베이스라인이 만든 춤들 사이의 비교가 이뤄질 수 있도록 동영상들을 페어링 하였다. 어떤 음악에서 만든 춤들을 어떤 순서로 페어링해 유저들에게 보여줬는지에 대한 정보는 서플멘터리 매트리얼에서 더 찾아볼 수 있다. 우리의 유저스터디 방법론은 LI의 워크를 참고하였음을 알아달라.

% 테이블 1의 결과의 Lose Rate는 베이스라인의 경우 제안하는 방법론과 비교당했을때 몇퍼센트의 확률로 졌는지, 제안하는 방법론의 경우 베이스라인과 비교당했을때 몇퍼센트의 확률로 졌는지, 를 나타낸다. 실험 결과, 제안하는 방법론에 리워드 모델이 있는 경우가 다른 베이스라인들보다 더 선호되는 것을 확인할 수 있었다. 특히, 더 복잡한 configuration을 가진 에이전트의 UR5의 경우, 제안하는 방법론이 더 압도적으로 편차를 보이며 선호 받는 다는 것을 확인할 수 있었다. 모션-비트 코릴레이션이 베이스라인들보다 상대적으로 낮음에도 불구하고 사람들로 선호받는다는 사실은 흥미로운데, 이는 '춤'이라는 것의 퀄리티를 평가하는 요소가 꼭 모션이 비트와 잘 맞는지만을 따지는 것이 아니기 때문이라고 생각되며, 더 자세한 내용은 다음 섹션 리미테이션에서 디스커스 할 것이다.

\section{Limitation}
\paragraph{Main Hypothesis}
As introduced in our abstract, our main hypothesis is \textit{``Dance is a motion that forms a visual rhythm from music, where the visual rhythm can be perceived from an optical flow"}. 
However, this hypothesis can arise two issues. 
First, since we assume that visual rhythm can be represented with optical flow, which does not contain 3D information such as depth, detailed motion information can be lost. 
Therefore, it will be necessary to devise a new modality that can represent the visual rhythm in a 3D space. 
Second, it is not only the visual rhythm synchronized with music that makes humans evaluate a certain motion as a `good dance'. This can be inferred from Table~\ref{tab:stats}, which shows that the proposed framework is most preferred by human subjects even though its motion-music correlation scores are sometimes lower than other baselines.
Regarding this, designing a new reward model that can consider various factors such as how the dancer effectively uses the given stage, and how much the dance is well-structured, would be necessary.

\paragraph{Unstructured Choreography}
In general, human dance is a sequence of basic primitive movements, and it is possible to record a choreography as a document based on a dance notation. In addition, human dance often repeats key movements corresponding to the main hook of the music, to catch the eye. However, the dance movement generated by the proposed framework is not structured like this. Rather, it is closer to an improvised dance on the spot. In this respect, existing works \cite{bi2018real} that bring out the primitive motions from the predefined motion library can be recognized as more effective in generating structured dance.
However, note that our objective is to present a framework that can train non-humanoid dancers only with human videos, without such a predefined motion library. 
To improve our method of generating structured dance, building a primitive dance motion library for non-humanoid agents based on human videos can be considered, which is one of our future works.

\section{Conclusion}
In this paper, we propose a framework for teaching any kind of non-humanoid agents how to dance based on human dance videos. To solve this problem, we first set our hypothesis, which defines dance as a motion forming a visual rhythm from music, where the visual rhythm can be perceived from an optical flow. 
Based on this, we employ a contrastive learning strategy to train a reward model which can understand the relationship between the optical flow (visual rhythm) and music from human dance videos. With this reward model, it is possible to train any kind of non-humanoid agent to dance by giving a higher reward if its motion creates an optical flow that is more correlated to the given music feature.
Experiment results show that our non-humanoid dancers can create a motion that is properly aligned with the music. Although the scores of motion-music correlation are sometimes relatively lower than the baselines, the user study result shows that the dance moves from the proposed framework are preferred by humans. 
To the best of our knowledge, our framework is the first to suggest training non-humanoid dancers from human dance videos. 
However, there is still room for improvement, such as devising a modality that can represent the visual rhythm in 3D space, or proposing a way to create more structured choreography.

\newpage

\section{Supplementary Material}
\subsection{User Study}
\subsubsection{Procedure}
In our user study, we asked participants to compare the dance videos made by our proposed framework, with dance videos made by other baselines. Let $A$ be the dance video from the proposed framework, and $B$ be the dance video from the baseline. We let participants watch $A$ and $B$ in random order (i.e., $A\rightarrow B$ or $B\rightarrow A$). After that, we showed two videos at once so that participants could have another chance to compare $A$ and $B$ at the same time. All videos were 30 seconds long.
Finally, we asked users ``Which non-humanoid agent dances better to the given music?". Users freely evaluated which agent danced better, as they felt. The median value of the participation time was around 17 minutes, and we rewarded the participants based on a pay rate of 8.98 pounds/hour. Note that the way of our user study referenced the work in \cite{li2021ai}

In total, nine questions were answered. Table I shows details about questions and results. Numbers in this table denote the vote results from 50 participants. Here, w/ RM denotes our framework with the reward model, w/o RM denotes our framework without the reward model, and BPM denotes the BPM-based Control baseline. A pair of methods were evaluated on three songs, which are two songs from AIST database and one pop music that is not included in AIST database.
In Table I, pop music \#1 was `Sincerity Is Scary' by The 1975, pop music \#2 was `Attention' by New Jeans, and pop music \#3 was `We will rock you' by Queen.
% In the case of UR5 agent, the proposed framework with the reward model was compared three times with BPM-based Control, and three times compared with the framework without the reward model. 
% In the case of Cartpole, the proposed framework with the reward model was compared three times with the framework without the reward model. 
% BPM-based Control was not considered as a baseline for CartPole, since it makes the CartPole agent go out of the screen so that the episode ends.

The result in Table I shows that the proposed framework with the reward model was preferred over the baselines in all cases. However, when the proposed framework with the reward model is compared with the framework without the reward model based on pop music and UR5 simulator, there is no significant difference in preference. 
Yet, it is not plausible to conclude that there is not much difference between the two models when unseen music was given as an input.
For better analysis, it would be better to conduct the user study based on more participants and more diverse songs. 
% However, it was not possible due to time and financial issues. If we can spare more time and financial budget, it would be better to conduct a more in-depth user study and analysis. 

\subsubsection{Participants}
We recruited a total of 55 participants through a site called Prolific\footnote{\url{https://www.prolific.co/}}. Among them, five people who completed the study within a much shorter time ( < 7 minutes) than our expectation ( $\sim$ 15 minutes) were excluded from the results and rewards. 
Before start asking a question, we also conducted a pre-survey on participants, regarding (1) whether they had any experience in dancing, (2) how often they watch dance videos. The result from this pre-survey is shown in Figure I. 
It is shown that about 52\% of the participants had the experience of dancing, and about 44\% of the participants frequently watch dance videos, weekly or daily.

\subsection{Agents and Simulators}
For our experiment, we used CartPole agent from Gym\footnote{\url{https://gymnasium.farama.org/}}, and UR5 agent from RoboSuite\footnote{\url{https://robosuite.ai/}}. In the case of CartPole, other works in reinforcement learning usually end the episode when the pole fails to stand upright. But in our case, since the goal is not about setting the pole vertically, we excluded that episode-ending condition. In the case of UR5 agent, considering the gripper attached to the robot arm, this agent has seven degrees of freedom. 
To control the robot, the joint position controller provided by RoboSuite was used. However, if the joint torque frequently reaches the maximum value (1.0), problems may occur when deploying it to actual hardware. Therefore, the obtained joint torque value was used after clipping the value between -0.9 and 0.9.

\begin{table}[]
\centering
\begin{tabular}{|cccccc|ccc|}
\hline
\multicolumn{6}{|c|}{UR5}                                                                                                                                                                                                                                      & \multicolumn{3}{c|}{CartPole}                                                                                        \\ \hline \hline
\multicolumn{1}{|c|}{Music}                                                     & \multicolumn{1}{c|}{w/ RM} & \multicolumn{1}{c||}{BPM} & \multicolumn{1}{c|}{Music}                                                     & \multicolumn{1}{c|}{w/ RM} & w/o RM & \multicolumn{1}{||c|}{Music}                                                     & \multicolumn{1}{c|}{w/ RM} & w/o RM \\ \hline
\multicolumn{1}{|c|}{mMH1}                                                      & \multicolumn{1}{c|}{33}    & \multicolumn{1}{c||}{17}  & \multicolumn{1}{c|}{mBR1}                                                      & \multicolumn{1}{c|}{40}    & 10     & \multicolumn{1}{||c|}{mJB1}                                                      & \multicolumn{1}{c|}{43}    & 7      \\ \hline
\multicolumn{1}{|c|}{mPO1}                                                      & \multicolumn{1}{c|}{35}    & \multicolumn{1}{c||}{15}  & \multicolumn{1}{c|}{mJS1}                                                      & \multicolumn{1}{c|}{33}    & 17     & \multicolumn{1}{||c|}{mLH1}                                                      & \multicolumn{1}{c|}{42}    & 8      \\ \hline
\multicolumn{1}{|c|}{\begin{tabular}[c]{@{}c@{}}Pop\\ Music\\ \#1\end{tabular}} & \multicolumn{1}{c|}{27}    & \multicolumn{1}{c||}{23}  & \multicolumn{1}{c|}{\begin{tabular}[c]{@{}c@{}}Pop\\ Music\\ \#2\end{tabular}} & \multicolumn{1}{c|}{31}    & 19     & \multicolumn{1}{||c|}{\begin{tabular}[c]{@{}c@{}}Pop\\ Music\\ \#3\end{tabular}} & \multicolumn{1}{c|}{31}    & 19     \\ \hline
\end{tabular}
\caption{Number of votes for nine questions from 50 participants in our user study.}
\end{table}

\begin{figure}[t]
\centering
\centerline{\includegraphics[width=\columnwidth]{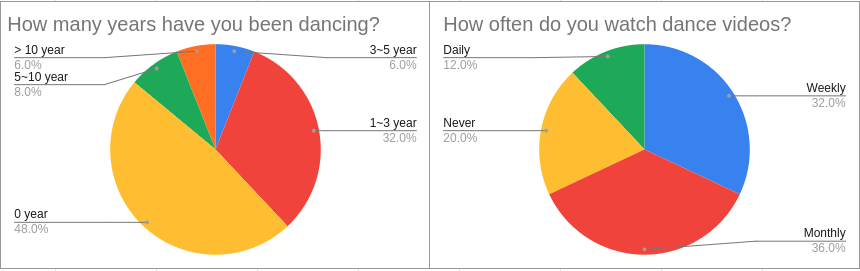}}
\caption{Results from pre-survey asking two questions about one's experience in dance.}
\label{fig:user_stat}
\end{figure}

\subsection{Reinforcement Learning Policy}
We used PPO algorithm\cite{ppo} to train the CartPole agent, and used SAC algorithm\cite{sac} to train UR5 agent. 
For both agents, the policy network's architecture was a multi-layered perceptron (MLP) with two hidden layers and a GELU activation layer in the middle. 
For the CartPole agent, the batch size was 2,048, the learning rate was 0.0001, and the rest of the hyperparameters were the default values provided by stable baseline3\footnote{\url{https://stable-baselines3.readthedocs.io/en/master/}}.
For the UR5 agent, the batch size was 2,048, the learning rate was 0.0001, and the buffer size was 3,000,000. The gradient update was conducted for every time step, after waiting 30,000 time steps in the beginning. For other hyperparameters, the default values provided by the stable baseline3 were also used.

When training a policy with these settings, the episode of CartPole started with its pole standing upright. The episode of UR5 started from a joint position that was randomly initialized. The maximum length of episodes was set to 1,000. 
For each episode, a random music piece from the AIST database was selected, and the starting time of the music was also randomly sampled.
\begin{ack}
Special thanks to Suji Moon for her thoughtful suggestions in shaping the title of this paper \smiley{}

% Do {\bf not} include this section in the anonymized submission, only in the final paper. You can use the \texttt{ack} environment provided in the style file to autmoatically hide this section in the anonymized submission.
\end{ack}

\newpage

{\small
\bibliographystyle{plainnat}
\bibliography{ref}
}

\end{document}